\documentclass{article} 
\usepackage{iclr2027_conference,times}

\usepackage{amsmath,amsfonts,bm}

\def\eqref#1{equation~\ref{#1}}

\def\1{\bm{1}}

\DeclareMathAlphabet{\mathsfit}{\encodingdefault}{\sfdefault}{m}{sl}
\SetMathAlphabet{\mathsfit}{bold}{\encodingdefault}{\sfdefault}{bx}{n}

\usepackage{hyperref}
\usepackage{url}

\usepackage{amssymb}
\usepackage{amsmath}
\usepackage{algorithm}
\usepackage{algpseudocode}
\usepackage{multirow}
\usepackage{graphicx}
\usepackage{subcaption}
\usepackage{xcolor}
\usepackage{booktabs}
\usepackage{caption}
\usepackage{wrapfig}
\usepackage{float}
\usepackage{makecell}

\title{ResQAdam: Residual-Preserving Adaptive Optimization for Memory-Efficient Quantization Aware Training}

\author{Liqun Ma$^{1}$\thanks{Work done during internship at AMD.}\ , Shengkun Tang$^{1}$, Liangyu Wang$^{2}$, Lin Zhao$^{3}$, Wei Luo$^{3}$, Xinjun Niu$^{3}$, \\
\textbf{Spandan Tiwari$^{3}$, Ashish Sirasao$^{3}$, Kun Zhang$^{1,4}$, Zhiqiang Shen$^{1}$} \\
$^{1}$Mohamed bin Zayed University of Artificial Intelligence (MBZUAI) \\
$^{2}$King Abdullah University of Science and Technology (KAUST) \\
$^{3}$Advanced Micro Devices, Inc. (AMD) \\
$^{4}$Carnegie Mellon University (CMU) \\
\small\texttt{\{liqun.ma, shengkun.tang, kun.zhang, zhiqiang.shen\}@mbzuai.ac.ae} \\
\small\texttt{liangyu.wang@kaust.edu.sa} \\
\small\texttt{\{lin.zhao, wei.luo, xinjun.niu, spandan.tiwari, ashish.sirasao\}@amd.com}
}

\iclrfinalcopy 
\begin{document}

\maketitle
\fancyhead{}
\renewcommand{\headrulewidth}{0pt}

\begin{abstract}

Quantization-aware training (QAT) is the most reliable way to obtain accurate low-bit large language models (LLMs), but training with AdamW keeps two dense FP32 states for every weight matrix. We find that low-rank memory-efficient optimizers such as GaLore and LDAdam degrade substantially in QAT. A considerable part of the QAT momentum lies outside its top-$r$ singular subspace, and projection-based methods discard this part at every step. We further observe that this residual remains useful when only its sign is kept. Based on these observations, we propose ResQAdam. It keeps a single BF16 momentum, applies Adam only to the $r\times r$ coordinates of the principal subspace, and adds a normalized sign update computed from the residual. We evaluate ResQAdam with two from-scratch QAT methods and three QAT fine-tuning pipelines, with weights quantized down to 2 bits. It reduces optimizer-state memory by 71\% to 74\% relative to AdamW, stays close to AdamW in most settings, and lowers training throughput by less than 2\%. Our code is available at \url{https://github.com/LiqunMa/ResQAdam}.

\end{abstract}

\section{Introduction}

Quantization-aware training (QAT) inserts the quantizer into the forward pass and updates latent full-precision weights through the straight-through estimator (STE)~\citep{bengio2013estimating}. 
At low bit widths it recovers much of the accuracy lost by post-training quantization (PTQ)~\citep{frantar2022gptq,xiao2023smoothquant}, both when fine-tuning a pretrained checkpoint~\citep{liu2024llmqat,chen2025efficientqat,liu2026paretoq} and when training from scratch~\citep{panferov2025quest,tabesh2025cage}.  
This accuracy comes at a memory cost. 
QAT keeps full-precision weights and gradients, and most recipes rely on AdamW~\citep{loshchilov2017decoupled}, which stores two FP32 moments and adds $8nm$ bytes for each $n\times m$ weight matrix.

\begin{figure}[t]
    \centering
    \includegraphics[width=0.7\columnwidth]{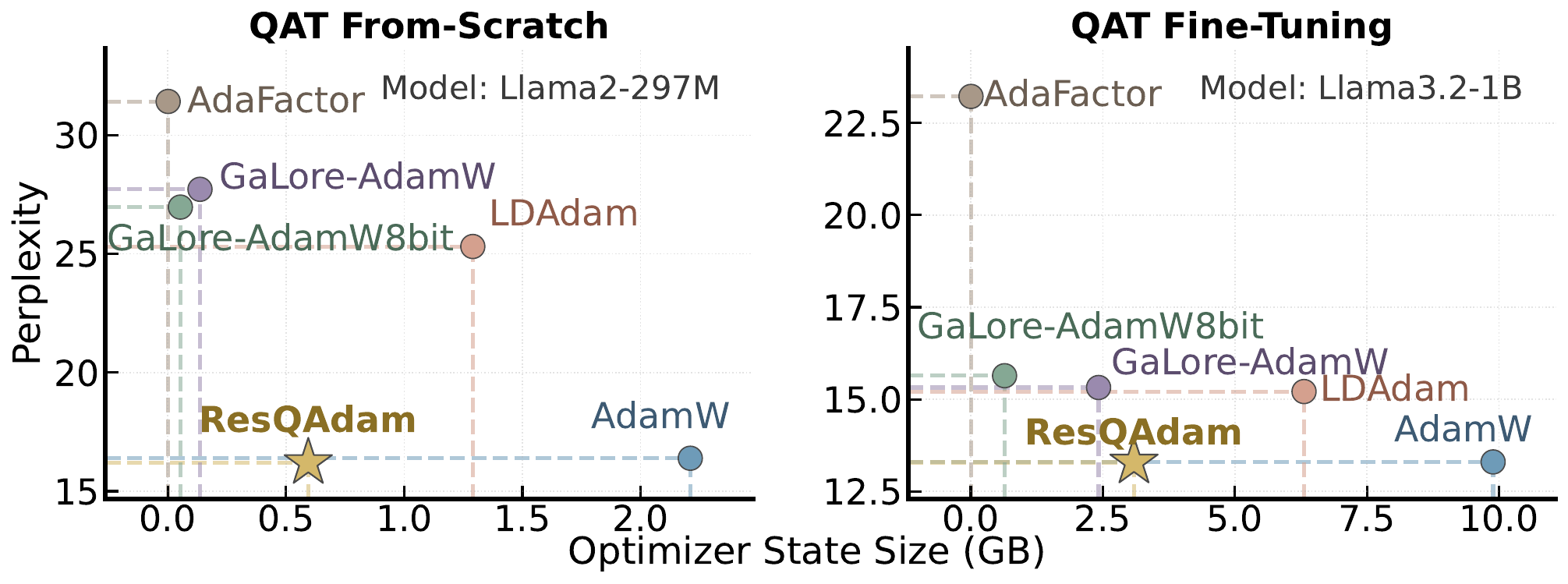}
    \vspace{-1em}
    \caption{WikiText-2 perplexity versus optimizer state memory for different optimizers under two QAT paradigms. The x-axis reports the total optimizer state size of the full model (in GB). All experiments use W2A16 quantization with group size 128, and projection-based optimizers use rank 64. For a fair comparison, the learning rate of each optimizer is tuned via grid search and the best result is reported.}
    \label{fig:memory_ppl}
    \vspace{-0.15in}
\end{figure}

A natural remedy is to borrow memory-efficient optimizers from full-precision training. 
GaLore~\citep{zhao2024galore} and LDAdam~\citep{robert2025ldadam} project the gradient onto a rank-$r$ subspace, keep Adam states only in that subspace, and map the update back to the full space. 
This works well when the optimization signal is concentrated in a few directions. Figure~\ref{fig:memory_ppl} shows that the same strategy does not carry over to QAT. 
With rank 64 and tuned learning rates, both methods are clearly worse than AdamW, in from-scratch training as well as in fine-tuning.

We hypothesize that this is a property of the QAT update itself. 
Under the STE, the gradient has to compensate for element-wise quantization errors, which are spread over individual entries rather than aligned with a few directions of the weight matrix. 
If these corrections are less concentrated in the top-$r$ subspace than the rest of the update, a projection-based optimizer discards them at every step, and no amount of tuning inside the subspace recovers them. 

To address this limitation, we decompose the QAT optimization signal according to its geometric structure.
We apply Singular Value Decomposition (SVD) to the dense global momentum, retaining the top-$r$ singular vectors to form the orthonormal bases $U_t \in \mathbb{R}^{n \times r}$ and $V_t \in \mathbb{R}^{m \times r}$.
These bases induce a principal matrix subspace $\mathcal{S}_t = \left\{ U_t A V_t^{\top} : A \in \mathbb{R}^{r \times r} \right\}$.
The momentum can then be exactly decomposed into its projection onto \(\mathcal{S}_t\) and an orthogonal residual lying in \(\mathcal{S}_t^{\perp}\).

The principal projection captures the dominant low-rank structure of the optimization trajectory, whereas the orthogonal residual retains dispersed signals outside the principal subspace.
Our empirical results show that these residual signals become particularly important under QAT.
Their magnitude matters little: keeping only their sign, with no second-moment state, recovers most of the lost quality (Section~\ref{sec:ablation}).

Motivated by this insight, we propose \textbf{ResQAdam} (\textbf{Res}idual-Preserving \textbf{Q}uantization-Aware \textbf{Adam}), a geometrically decoupled optimizer that applies distinct update rules based on subspace characteristics. 
It stores one dense momentum in BF16 and refreshes its top-$r$ singular vectors every $\tau$ steps. Adam runs on the $r\times r$ coordinates of the gradient in this subspace, and the residual of the momentum contributes a sign update whose norm matches the Adam update. 
Since no dense second moment is kept, persistent optimizer memory falls to roughly a quarter of AdamW, and the SVD cost is spread over $\tau$ steps.

By eliminating dense second-moment tracking, ResQAdam limits the persistent optimizer memory footprint to only the BF16 momentum and negligible low-rank components. 
For a weight matrix $W \in \mathbb{R}^{n \times m}$ with projection rank $r$, the persistent memory ratio compared to standard AdamW is:
$ 2nm + \mathcal{O}((n+m)r+r^2)/{8nm} \approx 25\% $.
The measured saving is 71\% to 74\%.
Furthermore, while the required SVD introduces computational overhead, refreshing the bases only once every few hundred steps effectively amortizes this cost, resulting in minimal training-speed degradation.

To evaluate our approach, we conduct extensive experiments across both from-scratch QAT and QAT fine-tuning paradigms. 
Across multiple LLM architectures, quantization configurations, and QAT methods, our empirical results demonstrate that ResQAdam closely tracks the optimization trajectory and largely preserves AdamW-level final performance across the evaluated settings. 
It achieves this by reducing around 75\% of AdamW's average memory footprint, while introducing negligible computational overhead for training.

\vspace{-0.1in}
\section{Related Work}
\vspace{-0.05in}
\textbf{From-scratch QAT.} Learnable quantizers such as PACT~\citep{choi2018pact} and LSQ~\citep{esser2019learned} were later scaled to LLMs by BitNet~\citep{wang2023bitnet} and BitNet b1.58~\citep{ma2024era}. Recent work improves the gradient estimate. QuEST~\citep{panferov2025quest} applies Hadamard transforms and down-weights gradients of entries with large quantization error, and CAGE~\citep{tabesh2025cage} adds a curvature-aware correction to the STE. Both quantize only the forward pass and keep gradients and optimizer states in high precision, which is the setting we target.

\textbf{QAT Fine-Tuning.}
A complementary paradigm adapts a converged full-precision checkpoint to recover its compressed representation. 
LLM-QAT~\citep{liu2024llmqat} performs data-free QAT by distilling from the pretrained model's own generations, jointly quantizing weights, activations, and the KV cache. 
EfficientQAT~\citep{chen2025efficientqat} reduces QAT cost via two stages: block-wise all-parameter training (Block-AP) for a strong low-bit initialization, followed by end-to-end training of only the quantization step sizes (E2E-QP). 
ParetoQ~\citep{liu2026paretoq} unifies 1-to-4-bit QAT under a single framework and identifies a sharp representation transition between 2 and 3 bits. 
Unlike parameter-efficient alternatives such as PEQA~\citep{kim2023memory}, which shrinks optimizer memory by training only the quantization scales, or QLoRA~\citep{dettmers2023qlora}, which fine-tunes high-precision adapters over a frozen quantized base, our method retains \emph{full-parameter} updates while still reducing optimizer memory.

\begin{wrapfigure}{r}{0.53\columnwidth}
    \vspace{-0.14in}
    \centering

    \hrule height 0.8pt

    \refstepcounter{algorithm}
    \label{alg:resqadam-main}
    
    \noindent
    \begin{minipage}[c][1.1\baselineskip][c]{\linewidth}
        \raggedright
        \textbf{Algorithm~\thealgorithm}  ResQAdam: Main Flow
    \end{minipage}
    
    \hrule height 0.4pt
    \vspace{4pt}

    \begin{minipage}{0.53\columnwidth}
    \small
    \begin{algorithmic}

\Statex \textbf{Require:} \textbf{Weights}
$W_0 \in \mathbb{R}^{n \times m}$,
\textbf{learning rate} $\eta$,
\textbf{weight decay} $\lambda$,
\textbf{rank} $r$,
\textbf{interval} $\tau$,
\textbf{Adam parameters} $\beta_1,\beta_2,\epsilon$

\Statex \textbf{Initialize:}
$M_0 \leftarrow 0$, \quad
$m_0^{\parallel},v_0^{\parallel}\leftarrow 0$

\For{$t=1,2,\dots$}
    \State \textbf{Obtain STE gradient} $G_t$

    \State $M_t \leftarrow
    \beta_1M_{t-1}+(1-\beta_1)G_t$

    \If{$t\bmod\tau=1$}
        \State $U_t,V_t\leftarrow
        \operatorname{SVD}(M_t,\operatorname{rank}=r)$
        \State Transport
        $m_{t-1}^{\parallel},v_{t-1}^{\parallel}$
        to new bases
    \EndIf

    \State $M_t^{\parallel}\leftarrow
    U_tU_t^\top M_tV_tV_t^\top$

    \State $M_t^{\perp}\leftarrow
    M_t-M_t^{\parallel}$

    \State $\Delta_t^{\parallel}\leftarrow
    \textsc{ProjectedAdamUpdate}
    (G_t,U_t,V_t)$

    \State $\Delta_t^{\perp}\leftarrow
    \textsc{ResidualSignUpdate}
    (M_t^{\perp},\Delta_t^{\parallel})$

    \State \textbf{Compute energy ratio}
    \Statex \hspace{\algorithmicindent}
    $\rho_t^{\parallel},\rho_t^{\perp}
    \leftarrow
    \textsc{CalcEnergyWeights}
    (M_t,M_t^{\parallel},M_t^{\perp})$

    \State \textbf{Fuse principal and residual directions}
    \Statex \hspace{\algorithmicindent}
    $\Delta_t\leftarrow
    \rho_t^{\parallel}\Delta_t^{\parallel}
    +\rho_t^{\perp}\Delta_t^{\perp}$

    \State $W_t\leftarrow
    (1-\eta\lambda)W_{t-1}-\eta\Delta_t$
\EndFor

\end{algorithmic}
    \end{minipage}

    \vspace{3pt}
    \hrule height 0.8pt
    \vspace{-10pt}
\end{wrapfigure}

\textbf{Low-rank Projection Optimization.}
Low-rank projection methods take a different route: GaLore~\citep{zhao2024galore} periodically computes an SVD of the gradient, runs Adam within the resulting compact subspace, and projects the update back, while LDAdam~\citep{robert2025ldadam} performs adaptive optimization inside dynamically shifting subspaces with a projection-aware transport rule and error feedback. 
These methods are most effective when optimization signals are strongly concentrated in a dominant low-rank subspace. 
Our experiments show that this assumption becomes substantially less effective under QAT: projection-only optimizers degrade markedly, while preserving the discarded residual restores much of the lost optimization quality.
While recent works like Fira~\citep{chen2026fira} and FRUGAL~\citep{zmushko2024frugal} also update the component outside the projected subspace with a state-free rule. 
ResQAdam differs in the residual is taken from the momentum rather than the current gradient, the bases are computed from the momentum, and the two updates are weighted by their momentum energy.

\section{ResQAdam}
\label{resq}
Based on the explicit orthogonal decomposition of optimization signals, we propose ResQAdam, a geometrically decoupled optimizer for QAT.
ResQAdam applies different update rules to two parts of the optimization space: a compact coordinate space in $\mathbb{R}^{r \times r}$, where Adam-style preconditioning is performed, and the full parameter space $\mathbb{R}^{n \times m}$, where the orthogonal residual provides a sign-based correction for dispersed quantization-compensation signals.
This decomposition substantially reduces the optimizer memory footprint while retaining the residual correction signals required for QAT.
The overall training flow is summarized in Algorithm~\ref{alg:resqadam-main}.

\subsection{Principal Subspace and Residual}
\label{sec:core_residual}
ResQAdam keeps no dense second moment. 
It keeps one dense momentum:
\begin{equation}
M_t = \beta_1 M_{t-1} + (1-\beta_1)G_t \in \mathbb{R}^{n \times m}.
\end{equation}
which is used to estimate the subspace and the residual direction but never to normalize the update. We store $M_t$ in BF16. 
Every $\tau$ steps, the top-$r$ left and right singular vectors of $M_t$ give $U_t\in\mathbb{R}^{n\times r}$ and $V_t\in\mathbb{R}^{m\times r}$. 
They span the principal subspace $\mathcal{S}_t=\{U_tAV_t^\top : A\in\mathbb{R}^{r\times r}\}$, which has dimension $r^2$. 
With the orthogonal projector $\Pi_t(X)=U_tU_t^\top XV_tV_t^\top$, the momentum splits into $M_t^{\parallel}=\Pi_t(M_t)$ and $M_t^{\perp}=M_t-M_t^{\parallel}$, which are orthogonal under the Frobenius inner product.

\subsection{Space-Specific Optimization Strategies}
\label{sec:subspace_opt}
\textbf{Adam in the principal subspace.} We project the gradient to $G_t^{\parallel}=U_t^\top G_tV_t$ and apply Adam to it, which gives preconditioned coordinates $\hat A_t^{\parallel}$ (Algorithm~\ref{alg:resqadam-core}). The update in parameter space is:
\begin{equation}
\Delta_t^{\parallel}=\alpha_r U_t\hat A_t^{\parallel}V_t^\top
\end{equation}
with $\alpha_r=\sqrt{nm}/r$. The back-projection preserves the Frobenius norm, so an $r\times r$ update with entries of order one has norm $O(r)$, whereas a dense update of the same entry scale has norm $O(\sqrt{nm})$. The factor $\alpha_r$ removes this dependence on $r$.

\textbf{Sign update from the residual.} A second moment for $M_t^{\perp}$ would bring back dense memory. We use only its sign and rescale it to the norm of the principal update:
\begin{equation}
\Delta_t^{\perp}=\|\Delta_t^{\parallel}\|_F\,\mathrm{sign}(M_t^{\perp})/\|\mathrm{sign}(M_t^{\perp})\|_F
\end{equation}
Because the sign acts element-wise, $\Delta_t^{\perp}$ is in general not orthogonal to $\mathcal{S}_t$. We therefore view it as a direction derived from the residual rather than an update restricted to $\mathcal{S}_t^{\perp}$.

\begin{wrapfigure}{r}{0.48\columnwidth}
    \vspace{-10pt}
    \centering

    \hrule height 0.8pt

    \refstepcounter{algorithm}
    \label{alg:resqadam-core}

    \noindent
    \begin{minipage}[c][1.1\baselineskip][c]{\linewidth}
        \raggedright
        \textbf{Algorithm~\thealgorithm}  Projected Adam Update
    \end{minipage}

    \hrule height 0.4pt
    \vspace{4pt}

    \begin{minipage}{0.53\columnwidth}
    \small
    \begin{algorithmic}

\State \textbf{Require:} $G_t$, $U_t,V_t$, $m_{t-1}^{\parallel},v_{t-1}^{\parallel}$

\State \textbf{Project gradient to the $r \times r$ coordinate space}
\Statex \hspace{\algorithmicindent} $G_t^{\parallel} \leftarrow U_t^\top G_t V_t$

\State $m_t^{\parallel} \leftarrow
\beta_1 m_{t-1}^{\parallel} + (1-\beta_1)G_t^{\parallel}$

\State $v_t^{\parallel} \leftarrow
\beta_2 v_{t-1}^{\parallel}
+ (1-\beta_2)G_t^{\parallel}\odot G_t^{\parallel}$

\State $\hat{m}_t^{\parallel},\hat{v}_t^{\parallel}
\leftarrow \text{BiasCorrect}(m_t^{\parallel},v_t^{\parallel})$

\State \textbf{Adam preconditioning in coordinate space}
\Statex \hspace{\algorithmicindent} $\hat{A}_t^{\parallel} \leftarrow
\hat{m}_t^{\parallel}/(\sqrt{\hat{v}_t^{\parallel}}+\epsilon)$

\State $\alpha_r \leftarrow \sqrt{nm}/r$

\State \textbf{Map update back to full space}
\Statex \hspace{\algorithmicindent} $\Delta_t^{\parallel}=\alpha_r U_t \hat{A}_t^{\parallel} V_t^\top$

\State \textbf{Return:} $\Delta_t^{\parallel}$

\end{algorithmic}
    \end{minipage}

    \vspace{4pt}
    \hrule height 0.8pt

    \refstepcounter{algorithm}
    \label{alg:resqadam-residual}

    \noindent
    \begin{minipage}[c][1.1\baselineskip][c]{\linewidth}
        \raggedright
        \textbf{Algorithm~\thealgorithm}  residual-sign Update
    \end{minipage}

    \hrule height 0.4pt
    \vspace{4pt}

    \begin{minipage}{0.53\columnwidth}
    \small
    \begin{algorithmic}

\State \textbf{Require:} $M_t^{\perp},\ \Delta_t^{\parallel}$

\State \textbf{Use only the residual direction}
\Statex \hspace{\algorithmicindent} $D_t^{\perp} \leftarrow \operatorname{sign}(M_t^{\perp})$

\State \textbf{Align residual magnitude with the}
\State \textbf{principal update}
\Statex \hspace{\algorithmicindent} $\Delta_t^{\perp} \leftarrow
\|\Delta_t^{\parallel}\|_F \cdot \frac{D_t^{\perp}}{\|D_t^{\perp}\|_F+\epsilon}$

\State \textbf{Return:} $\Delta_t^{\perp}$

\end{algorithmic}
    \end{minipage}

    \vspace{3pt}
    \hrule height 0.8pt
    \vspace{-2\baselineskip}
\end{wrapfigure}

\subsection{Energy-Aware Geometric Fusion}
\label{sec:energy_fusion}
The principal and residual components capture complementary structures of the optimization signal. To combine them in the full parameter space, we adapt their contributions according to the relative momentum energy carried by the two components.
Because $M_t^{\parallel}$ and $M_t^{\perp}$ form an orthogonal decomposition of the momentum $M_t$, their squared Frobenius norms satisfy:
\begin{equation}
\|M_t\|_F^2 = \|M_t^{\parallel}\|_F^2 + \|M_t^{\perp}\|_F^2 .
\end{equation}
We therefore define the energy ratio as:
\begin{equation}
\rho_t^{\parallel} = \frac{\|M_t^{\parallel}\|_F}{\|M_t\|_F+\epsilon}, \qquad
\rho_t^{\perp} = \frac{\|M_t^{\perp}\|_F}{\|M_t\|_F+\epsilon} .
\end{equation}

As described in Section~\ref{sec:subspace_opt}, the residual update is normalized to match the Frobenius norm of the principal update. We then combine the two directions using the momentum-derived energy ratio:
\begin{equation}
\Delta_t = \rho_t^{\parallel} \Delta_t^{\parallel} + \rho_t^{\perp} \Delta_t^{\perp} 
\end{equation}
Although $M_t^{\parallel}$ and $M_t^{\perp}$ are orthogonal by construction, the element-wise sign transformation does not preserve this orthogonality. 
Therefore, the resulting update directions $\Delta_t^{\parallel}$ and $\Delta_t^{\perp}$ are generally not orthogonal. 
We consequently interpret $\rho_t^{\parallel}$ and $\rho_t^{\perp}$ as energy-derived balancing coefficients rather than as weights that preserve an exact orthogonal norm decomposition.

Finally, applying weight decay $\lambda$ and learning rate $\eta$, the parameter update at step $t$ is:

\begin{equation}
W_{t} = (1-\eta\lambda)W_{t-1} - \eta\Delta_t
\end{equation}

\subsection{Subspace Refresh and Memory}
\label{sec:refresh_complexity}
The SVD is computed only every $\tau$ steps to amortize the computational cost. 
After each refresh, the principal states $m^{\parallel}$ and $v^{\parallel}$ are rewritten in the new basis with the transport rule of LDAdam~\citep{robert2025ldadam} (Appendix~\ref{app:state_transport}). 
Only these $r\times r$ states need transport, because $M_t$ and its residual are defined in the original parameter space. 

By storing the momentum $M _t$
in BF16 and confining adaptive states to the compact coordinate representation, the dominant persistent-state cost of ResQAdam is $2nm$ bytes, with an additional $O((n+m)r+r^2)$ overhead for the bases and principal states.
For $r\ll\min(n,m)$ this approaches 25\% of AdamW. The measured reduction is 71\% to 74\% (Section~\ref{sec:experiments}), mainly because the bases are not negligible for the smaller matrices.

\begin{figure}[t]
    \centering
    \includegraphics[width=0.85\columnwidth]{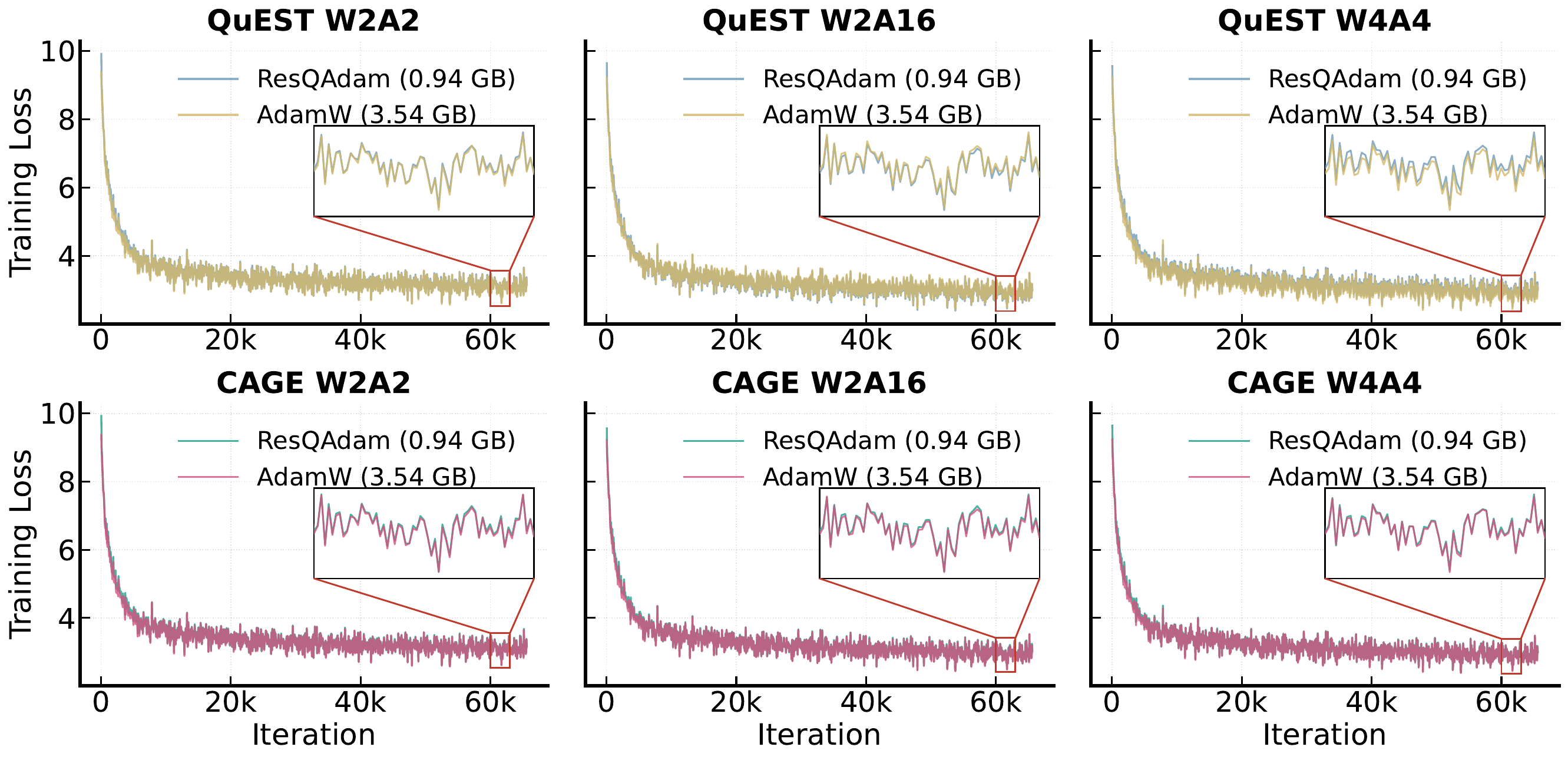}
    \vspace{-1em}
    \caption{Training loss trajectories of ResQAdam and AdamW during from-scratch QAT under various quantization configurations on the 430M-parameter LLM. The values following the optimizer names in the legend indicate the average memory footprint of the optimizer states.}
    \label{fig:loss}
    \vspace{-12px}
\end{figure}

\section{Experiments}
\label{sec:experiments}
We compare ResQAdam with AdamW in from-scratch QAT and QAT fine-tuning, in terms of training loss, final perplexity and accuracy, and optimizer-state memory.

\subsection{Experimental Setup}
\label{sec:exp_setup}
\textbf{From-scratch QAT.} 
We consider QuEST~\citep{panferov2025quest} and CAGE~\citep{tabesh2025cage}, which represent two recent low-precision training frameworks with distinct gradient-estimation and optimization strategies. We pretrain Llama-style models at the 200M and 430M scales under W2A2, W2A16, and W4A4 settings.

\textbf{QAT fine-tuning.} 
We use Basic QAT (end-to-end next-token loss with LSQ scales on 2B RedPajama tokens), LLM-QAT~\citep{liu2024llmqat} (data-free distillation), and the Block-AP stage of EfficientQAT~\citep{chen2025efficientqat}, on Llama 3.2-1B, Qwen3 4B-Base and Llama 2-7B under multiple quantization configurations.. Appendix~\ref{app:experiments_cfg} gives the details.

\begin{table}[b]
  \centering
  \scriptsize
  \vspace{-10px}
  \caption{Language modeling performance comparison across different from-scratch QAT methods. Values in
each cell are formatted as AdamW / ResQAdam (Ours). \textit{Q-Cfg} denotes the quantizer configuration, where \textit{W$x$} and \textit{A$x$} indicate $x$-bit weight and activation quantization, respectively. \textit{Opt. State Size} reports average optimizer state size for each optimizer during the training, computed only over non-embedding and non-LM-head parameters.}
\vspace{-1em}
  \label{tab:pretrain}
  \resizebox{.9\textwidth}{!}{
\begin{tabular}{c|c|c|c|c|c|c}
\toprule
Methods                & Size                  & Q-Cfg & \makecell{WikiText-2 PPL \\ (AdamW / Ours)}                           & \makecell{C4 PPL\\ (AdamW / Ours)}                                  & \makecell{Avg. PPL\\ (AdamW / Ours)}                                & \makecell{Opt. State Size   \\ (AdamW / Ours)}                \\ \hline\hline
\multirow{3}{*}{QuEST} & \multirow{6}{*}{200M} & W2A2  & 50.56 / 50.98 & 26.99 / 27.41 & 38.78 / 39.20 & \multirow{6}{*}{\makecell{1.63 / \textbf{0.44} GB \\ $\downarrow$ 73.0\%}} \\
                       &                       & W2A16 & 44.04 / 42.80 & 22.92 / 23.41 & 33.48 / 33.11 &                                   \\
                       &                       & W4A4  & 40.87 / 41.93 & 22.06 / 22.46 & 31.46 / 32.20 &                                   \\ \cline{1-1} \cline{3-6}
\multirow{3}{*}{CAGE}  &                       & W2A2  & 51.15 / 52.32 & 27.19 / 27.65 & 39.17 / 39.99 &                                   \\
                       &                       & W2A16 & 44.80 / 44.44 & 23.18 / 23.73 & 33.99 / 34.09 &                                   \\
                       &                       & W4A4  & 41.22 / 42.68 & 22.21 / 22.61 & 31.71 / 32.66 &                                   \\ \hline\hline
\multirow{3}{*}{QuEST} & \multirow{6}{*}{430M} & W2A2  & 39.12 / 41.48 & 21.69 / 22.11 & 30.40 / 31.80 & \multirow{6}{*}{\makecell{3.54 / \textbf{0.94} GB\\ $\downarrow$ 73.4\%}} \\
                       &                       & W2A16 & 34.74 / 34.66 & 18.79 / 19.18 & 26.77 / 26.92 &                                   \\
                       &                       & W4A4  & 34.00 / 34.13 & 18.16 / 18.39 & 26.08 / 26.26 &                                   \\ \cline{1-1} \cline{3-6}
\multirow{3}{*}{CAGE}  &                       & W2A2  & 40.49 / 41.24 & 21.81 / 22.25 & 31.15 / 31.75 &                                   \\
                       &                       & W2A16 & 35.93 / 35.29 & 19.08 / 19.42 & 27.50 / 27.36 &                                   \\
                       &                       & W4A4  & 34.55 / 34.13 & 18.30 / 18.57 & 26.43 / 26.35 &                                   \\ \bottomrule
\end{tabular}
}
\end{table}

\textbf{Hyperparameters and evaluation.} 
For from-scratch QAT, ResQAdam uses rank $r=32$ and refresh interval $\tau=400$.
As an implementation detail of Algorithm~\ref{alg:resqadam-main}, the bases are initialized from the first gradient and the periodic refresh is suspended during the first 10\% of training steps, since the momentum of a randomly initialized model changes rapidly and early refreshes only add transport noise; after this warm-up phase, refreshes follow the schedule of Algorithm~\ref{alg:resqadam-main}. For QAT fine-tuning, we set $r=64$ and $\tau=200$ and apply no suspension, because the pretrained checkpoint already provides a stable momentum.
For QAT fine-tuning, we set $r=64$ and $\tau=200$ to accommodate larger model capacities.
All other training hyperparameters are kept strictly consistent with the experimental codebases of the respective QAT methods.

We report perplexity on WikiText-2 and C4 to evaluate models pretrained with QAT.
For QAT fine-tuning, in addition to perplexity, we evaluate zero-shot accuracy on several standard benchmarks: HellaSwag~\citep{zellers2019hellaswag}, PIQA~\citep{bisk2020piqa}, ARC~\citep{clark2018think}, WinoGrande~\citep{sakaguchi2021winogrande}, and MMLU~\citep{hendryckstest2021}.

\begin{table}[]

\centering
\scriptsize
\caption{Performance and memory footprint comparison between ResQAdam and standard AdamW across diverse QAT fine-tuning frameworks.  Values in
each cell are formatted as AdamW / ResQAdam (Ours). \textit{Q-Cfg} denotes the quantizer configuration: \textit{W$x$} and \textit{A$x$} indicate $x$-bit weight and activation quantization, respectively; \textit{G128} represents group-wise quantization with a group size of 128; and \textit{C} denotes channel-wise quantization. \textit{Opt. State Size} reports average optimizer state size of AdamW / ResQAdam during the training , computed only over non-embedding and non-LM-head parameters.}
\vspace{-1em}
\label{tab:main_results}
\begin{tabular}{c|c|c|c|c|c}
\toprule
Method                        & Model                      & Q-Cfg     & \makecell{Avg. Task Acc. \\ (AdamW / Ours)}  & \makecell{Avg. PPL \\ (AdamW / Ours)} & \makecell{Opt. State Size \\ (AdamW / Ours)} \\ \hline\hline
\multirow{2}{*}{Basic QAT}          & \multirow{8}{*}{Llama 3.2-1B} & W2A16G128 & 47.56 / 47.36                                                               & 16.92 / 17.73                                                         & \multirow{2}{*}{\makecell{7.78 / \textbf{2.16} GB \\$\downarrow72.2\%$}}   \\
                              &                              & W4A4G128  & 51.29 / 51.29                                                               & 15.12 / 14.91                                                         &                    \\ \cline{1-1} \cline{3-6}
\multirow{3}{*}{LLM-QAT}      &                              & W2A16C    & 38.26 / 39.37                                                               & 43.73 / 35.05                                                         &\multirow{3}{*}{\makecell{7.78 / \textbf{2.16} GB \\$\downarrow72.2\%$}}                    \\
                              &                              & W4A16C    & 51.16 / 52.00                                                               & 14.03 / 14.30                                                         &                     \\
                              &                              & W4A4C     & 38.70 / 38.53                                                               & 40.91 / 39.21                                                         &                     \\ \cline{1-1} \cline{3-6}
\multirow{3}{*}{EfficientQAT (Block-AP)} &                              & W2A16G128 & 35.14 / 35.43                                                               & 351.78 / 379.52                                                       & \multirow{3}{*}{\makecell{494 / \textbf{141} MB/block \\ $\downarrow71.5\%$} }                    \\
                              &                              & W3A16G128 & 49.93 / 49.31                                                               & 16.73 / 16.79                                                         &                     \\
                              &                              & W4A16G128 & 54.43 / 54.56                                                               & 12.78 / 12.81                                                         &                    \\ \hline\hline
\multirow{2}{*}{Basic QAT}          & \multirow{8}{*}{Qwen3 4B-Base}    & W2A16G128 & 61.29 / 61.94                                                               & 12.76 / 12.88                                                         & \multirow{2}{*}{\makecell{29.07 / \textbf{7.78} GB \\ $\downarrow73.2\%$}}   \\
                              &                              & W4A4G128  & 68.21 / 67.91                                                               & 12.27 / 12.20                                                         &                     \\ \cline{1-1} \cline{3-6}
\multirow{3}{*}{LLM-QAT}      &                              & W2A16C    & 40.85 / 38.02                                                               & 33.84 / 46.72                                                         &  \multirow{3}{*}{\makecell{29.07 / \textbf{7.78} GB \\ $\downarrow73.2\%$}}                  \\
                              &                              & W4A16C    & 68.07 / 66.81                                                               & 11.68 / 11.93                                                         &                     \\
                              &                              & W4A4C     & 45.70 / 45.19                                                               & 35.46 / 31.32                                                         &                      \\ \cline{1-1} \cline{3-6}
\multirow{3}{*}{EfficientQAT (Block-AP)} &                              & W2A16G128 & 45.89 / 43.09                                                               & 34.92 / 51.25                                                         &  \multirow{3}{*}{\makecell{820 / \textbf{220} MB/block \\ $\downarrow72.0\%$}}                  \\
                              &                              & W3A16G128 & 66.69 / 66.88                                                               & 12.06 / 12.23                                                         &                     \\
                              &                              & W4A16G128 & 69.29 / 69.49                                                               & 10.88 / 10.94                                                         &                     \\ \hline\hline
\multirow{3}{*}{LLM-QAT}      & \multirow{6}{*}{Llama 2-7B}   & W2A16C    & 43.03 / 42.31                                                               & 12.25  / 13.93                                                        & \multirow{3}{*}{\makecell{51.81 / \textbf{13.28} GB \\$\downarrow 74.4\%$}}    \\
                              &                              & W4A16C    & 63.11 / 61.68                                                               & 6.82 / 6.89                                                           &                     \\
                              &                              & W4A4C     & 47.12 / 46.87                                                               & 14.12 / 12.58                                                         &                     \\ \cline{1-1} \cline{3-6}
\multirow{3}{*}{EfficientQAT (Block-AP)} &                              & W2A16G128 & 49.10 / 47.95                                                               & 24.61 / 12.91                                                         &\multirow{3}{*}{\makecell{1644 / \textbf{451} MB/block \\$\downarrow72.6\%$}}                  \\
                              &                              & W3A16G128 & 61.37 / 61.28                                                               & 6.98 / 6.99                                                           &                     \\
                              &                              & W4A16G128 & 63.77 / 63.76                                                               & 6.49 / 6.49                                                           &                    \\ \bottomrule
\end{tabular}
\vspace{-12px}
\end{table}

\subsection{Main Results on From-Scratch QAT}
\label{sec:results_from_scratch}

\textbf{Optimization Trajectory and Convergence Alignment.}
As illustrated in Figure~\ref{fig:loss}, ResQAdam closely tracks the training-loss trajectory of the dense AdamW baseline. 
Across all evaluated bit-width configurations and both  frameworks, the training loss curves of ResQAdam closely overlap with the baseline from initialization through to the final iterations.
The zoomed-in callouts at the end of the training horizons further highlight that ResQAdam maintains stable convergence without divergence. 
This confirms that our geometry-decoupled and space-specific update strategies sufficiently capture the quantization errors to stabilize the low-bit pretraining process. Curves for the 200M model are in Appendix~\ref{app:200m}.

\textbf{Language Modeling Quality.}
Table~\ref{tab:pretrain} reports the final validation perplexity across model scales and quantization settings. 
For both QuEST and CAGE, ResQAdam achieves perplexities close to full-state AdamW, with several cases where ResQAdam is slightly better. 
These results show that ResQAdam largely preserves AdamW-level language modeling quality in from-scratch QAT while substantially reducing optimizer-state memory.

\subsection{Main Results on QAT Fine-Tuning}
\label{sec:results_fine_tuning}
Table~\ref{tab:main_results} compares ResQAdam with AdamW across the three fine-tuning recipes, which differ in data source, training objective and optimization granularity. 
Across these settings, ResQAdam generally achieves performance close to AdamW in both perplexity and zero-shot accuracy, while reducing optimizer-state memory by 71\% to 74\%. 
Although some aggressive W2A16 block-wise settings show larger metric fluctuations, the overall results indicate that ResQAdam can serve as a memory-efficient alternative to AdamW for diverse QAT fine-tuning pipelines.

\section{Analysis}
\label{sec:analysis}

\subsection{Component Ablation and Subspace Design}
\label{sec:ablation}
We ablate the main components of ResQAdam on a 297M model under W2A2, W2A16, and W4A4 from-scratch QAT. 
To keep the ablation aligned with the goal of memory-efficient QAT, we restrict the comparison to lightweight variants that do not introduce substantial extra computation or optimizer-state memory. 
Variants whose added cost would eliminate the memory or throughput advantage over AdamW are excluded, since they would not be practical alternatives to AdamW.

\noindent \textit{\textbf{Principal Update Only:}} Uses only the Adam-preconditioned update in the principal subspace, i.e., $\Delta_t=\Delta_t^{\parallel}$, and removes the sign correction from the orthogonal residual.
    
\noindent \textit{\textbf{Residual-Sign Only:}} Uses only the normalized sign direction derived from the orthogonal residual $M_t^{\perp}$, and removes the Adam-preconditioned principal update.
    
\noindent \textit{\textbf{Gradient-SVD Bases:}} Computes the projection bases $U_t,V_t$ from the instantaneous gradient $G_t$ instead of the momentum $M_t$, while keeping the remaining update rule unchanged.
    
\noindent \textit{\textbf{Global-Momentum Sign:}} Applies a sign update directly to the momentum $M_t$ without decomposing it into the principal component and the orthogonal residual.
    
\noindent \textit{\textbf{ResQAdam:}} Uses the full design: Adam-style preconditioning in the $r \times r$ coordinate representation, residual-sign correction from $M_t^{\perp}$, and momentum energy-weighting to fuse the two updates.

\begin{wrapfigure}{r}{0.45\linewidth}
    \centering
    \vspace{-0.8em}
    \includegraphics[width=\linewidth]{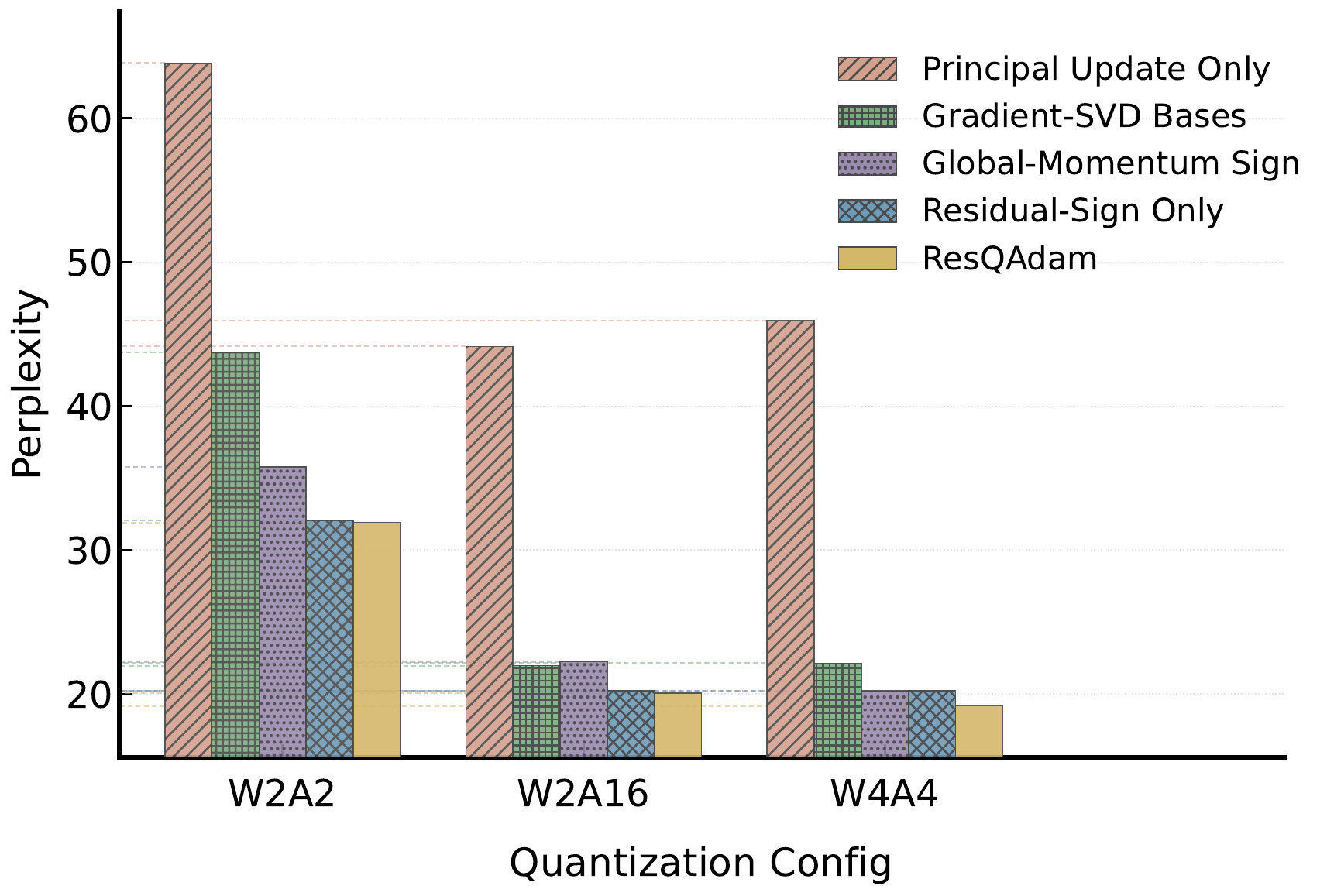}
    \vspace{-0.2in}
    \caption{Component ablation of ResQAdam across different quantization configurations. The y-axis reports the average perplexity on C4 and WikiText-2.}
    \label{fig:ablation_method}
    \vspace{-1.5em}
\end{wrapfigure}

Figure~\ref{fig:ablation_method} reports the average perplexity on C4 and WikiText-2. 
The Principal Update Only variant performs poorly across all quantization settings, showing that updates restricted to the principal subspace are not sufficient for QAT. 
The Residual-Sign Only variant performs much better, especially under W2A2 and W2A16, which indicates that the orthogonal residual contains useful correction signals. 
However, it is not consistently better than the full method, suggesting that the Adam-preconditioned principal update still provides important low-rank structural information.

The Gradient-SVD Bases variant underperforms ResQAdam, particularly under W2A2, validating the extraction of projection bases from the time-smoothed momentum rather than instantaneous gradients. 
Similarly, the Global-Momentum Sign variant falls short, demonstrating that applying a monolithic sign update negates the benefits of separating the principal component from the residual. 
Overall, the full ResQAdam consistently achieves the best perplexity across the evaluated settings, supporting the effectiveness of combining principal-subspace Adam preconditioning with a residual-derived sign correction.

\subsection{Hardware Efficiency and Overhead}
\label{sec:speed}

\begin{table}[t]
\scriptsize
\centering
\caption{Comparison of optimizer-state memory, total peak memory, and training throughput across different models. Values in each cell are formatted as \textit{AdamW / ResQAdam}, with relative changes in parentheses.}
\vspace{-0.1in}
\label{tab:speed}
\resizebox{.85\textwidth}{!}{
\begin{tabular}{lccc}
\toprule
\textit{AdamW / ResQAdam}    & \textbf{Llama 3.2-1B}           & \textbf{Qwen2.5-1.5B}         & \textbf{Qwen3 4B-Base}                \\ \midrule
\textbf{Opt. State Memory (GB)}       & 7.78 / 2.16 ($\downarrow$72.2\%)  & 10.48 / 3.01($\downarrow$71.3\%) & 29.07 / 7.78 ($\downarrow$73.2\%)   \\ \midrule
\textbf{Total Peak Memory (GB)} & 33.33 / 28.24 ($\downarrow$15.3\%)  &43.93 / 37.19 ($\downarrow$15.3\%)                      & 99.14 / 79.77 ($\downarrow$19.5\%) \\ \midrule
\textbf{Throughput (\# token/s)} & 29.36k / 28.94k ($\downarrow$1.4\%)  &18.31k / 18.19k ($\downarrow$0.6\%)                      & 9.64k / 9.56k ($\downarrow$0.8\%) \\ \bottomrule
\end{tabular}
}
\end{table}

To evaluate hardware efficiency, we compare ResQAdam against AdamW in terms of optimizer-state memory, total peak memory, and training throughput. 
Total peak memory is defined as the maximum GPU memory consumed during training, encompassing weights, activations, gradients, and optimizer states. 
All experiments utilize the Basic QAT framework with a W4A4G128 quantization configuration and a context length of 4096. Training is conducted on a single NVIDIA H200 GPU. 
The micro-batch size is set to 1 with 128 gradient accumulation steps.
ResQAdam uses a uniform projection rank of $r=64$. 
The reported optimizer-state memory strictly accounts for the transformer backbone and excludes the embedding and language modeling head layers.

As detailed in Table \ref{tab:speed}, ResQAdam consistently reduces the optimizer state footprint by over 71\% across 3 different LLMs. This efficiency directly translates to a 15.3\% to 19.5\% reduction in total peak memory. 
This substantial memory reduction incurs minimal computational overhead. 
The training throughput decreases by less than 2\% compared to standard AdamW in all evaluated cases. 
These results show that our amortized update strategy substantially alleviates the optimizer-state memory bottleneck of QAT while maintaining competitive training throughput.

\subsection{Comparison with Alternative Memory-Efficient Optimizers}
\begin{table}[t]
\scriptsize
\centering
\caption{Perplexity comparison of memory-efficient optimizers under diverse QAT paradigms and quantization configurations. Projection-based optimizers use rank 64. For a fair comparison, all optimizers are tuned with a grid search over learning rates, and the best result is reported.}
\label{tab:optimizer_ppl}
\vspace{-0.1in}
\begin{tabular}{c|c|ccccc}
\toprule
Training Paradigm \rule[-1.5ex]{0pt}{4ex}                & Q-Cfg      & ResQAdam & GaLore-AdamW & GaLore-AdamW8bit & LDAdam & AdaFactor \\ \hline
\multirow{2}{*}{From-Scratch QAT} & W2A16G128 \rule[-0.8ex]{0pt}{3ex}   & \textbf{20.05}    & 30.61       & 30.39           &  28.57  & 33.01     \\ \cline{2-7} 
                                  & W4A4G128 \rule[-0.8ex]{0pt}{3ex}   & \textbf{18.89}    & 32.93       & 32.45           & 28.23  & 32.84     \\ \hline \hline
\multirow{2}{*}{QAT Fine-Tuning}  & W2A16G128 \rule[-0.8ex]{0pt}{3ex}   & \textbf{16.89}    & 19.25       & 20.32           & 19.59  &  27.79     \\ \cline{2-7} 
                                  & W4A4G128 \rule[-0.8ex]{0pt}{3ex}   & 15.05    & 15.30       & 15.64           & 15.42  & \textbf{14.80}    \\ \bottomrule
\end{tabular}
\vspace{-10px}
\end{table}

We further compare ResQAdam with two families of memory-efficient optimizers. Low-rank projection methods such as GaLore~\citep{zhao2024galore} and LDAdam~\citep{robert2025ldadam} restrict adaptive states to a rank-$r$ subspace obtained from the gradient, with LDAdam adding state transport across changing subspaces and error feedback. Structural factorization methods such as AdaFactor~\citep{shazeer2018adafactor} keep the full parameter dimensionality but approximate the second moment by row and column statistics.

Table~\ref{tab:optimizer_ppl} reports the average perplexity on WikiText-2 and C4. In from-scratch QAT, every baseline is far behind ResQAdam in both quantization settings, by a margin much larger than the differences we observe between ResQAdam and AdamW in Table~\ref{tab:pretrain}. This is consistent with our hypothesis that confining the update to a low-rank subspace, or replacing the second moment by a factorized approximation, loses part of the correction that QAT relies on. In fine-tuning the picture depends on the bit width. At W2A16G128 the projection-based methods remain clearly behind, whereas at W4A4G128, where quantization error is small, all optimizers land within a narrow range and the choice of optimizer matters little.

\subsection{Sensitivity to $r$ and $\tau$}
\label{sec:sensitivity}
\begin{figure}[h]
    \centering
    \vspace{-0.15in}
    \begin{subfigure}{0.4\columnwidth}
        \centering
        \includegraphics[width=\columnwidth]{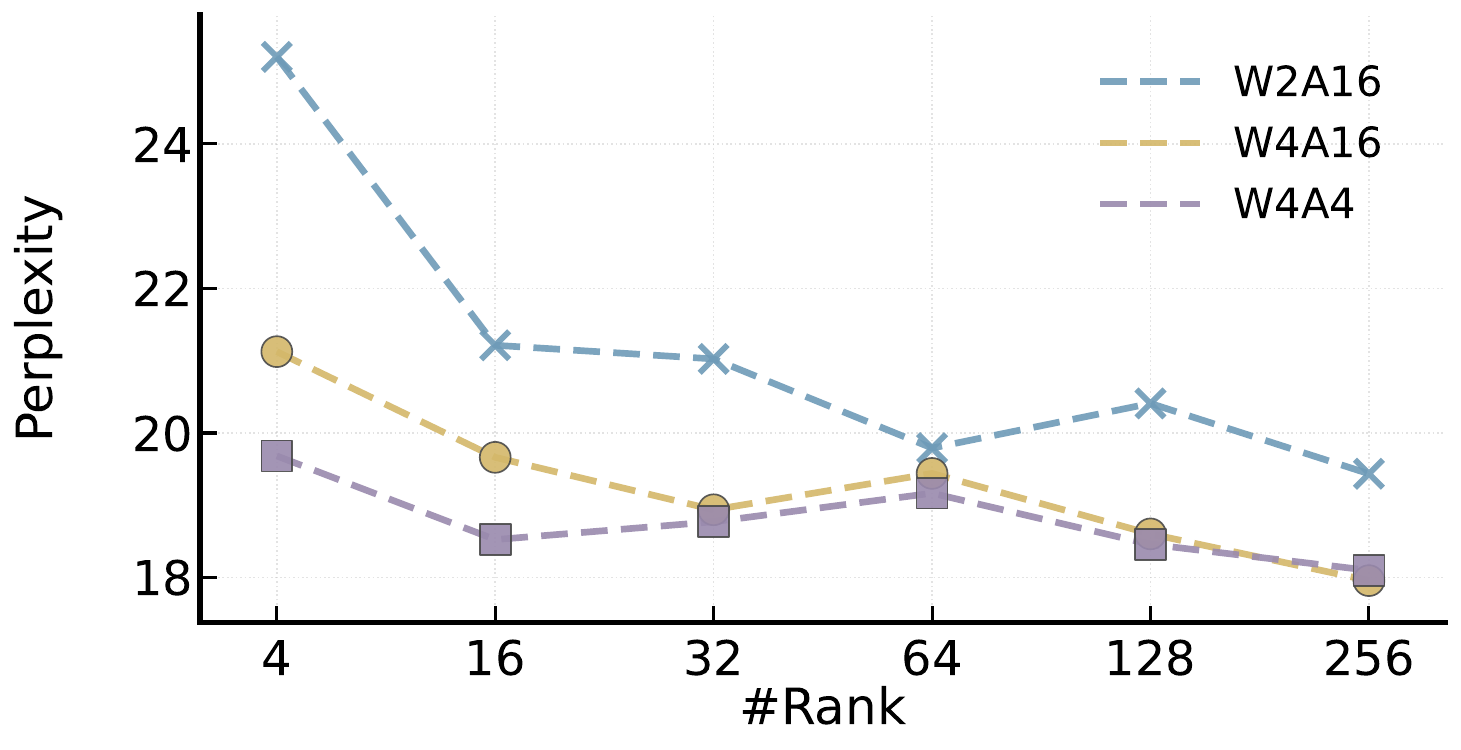}
        \caption{Rank $r$ ($\tau=400$).}
        \label{fig:ablation_r}
    \end{subfigure}
    \hspace{0.02\columnwidth}
    \begin{subfigure}{0.4\columnwidth}
        \centering
        \includegraphics[width=\columnwidth]{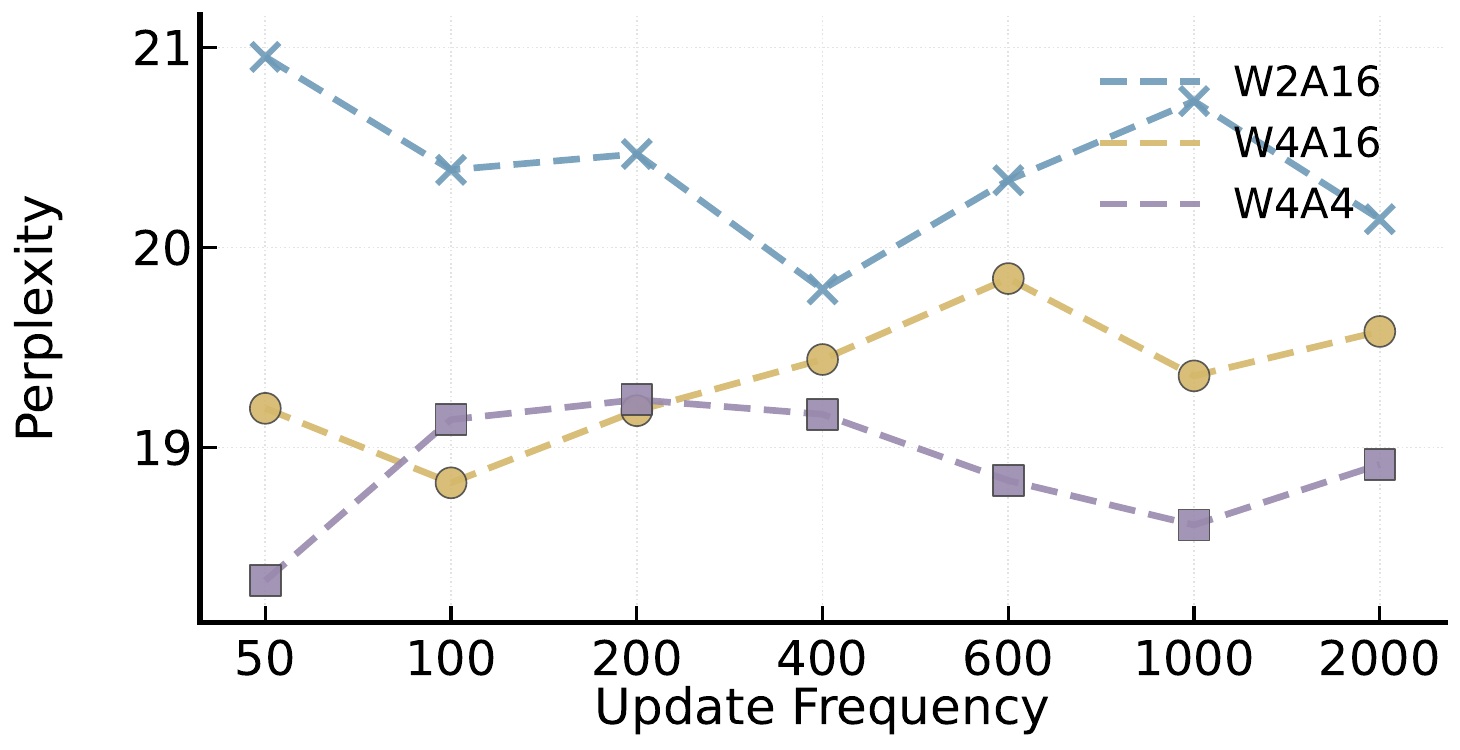}
        \caption{Refresh interval $\tau$ ($r=64$).}
        \label{fig:ablation_step}
    \end{subfigure}
    \vspace{-0.5em}
    \caption{Sensitivity of ResQAdam to the rank $r$ and the SVD refresh interval $\tau$ on a 297M model trained from scratch. The y-axis is the average of WikiText-2 and C4 perplexity.}
    \label{fig:group}
\end{figure}

The rank $r$ sets the dimension of the principal subspace, and the refresh interval $\tau$ sets how many steps a subspace is used before the bases are recomputed. We vary one of them while fixing the other at $r=64$ or $\tau=400$, using the 297M model trained from scratch with Basic QAT under W2A16, W4A16 and W4A4 (Figure~\ref{fig:group}).

\textbf{Rank.} A rank of 4 gives the highest perplexity in all three settings. The effect is largest at W2A16, where perplexity reaches 25.2 compared with 19.8 at $r=64$. For $r\geq 16$, perplexity no longer decreases monotonically with $r$. Within each setting it varies by at most 1.8 over $r\in\{16,32,64,128,256\}$, and $r=256$ gives the lowest value in all three settings. A very small subspace therefore hurts clearly, while above $r=16$ the effect of $r$ is small and not monotonic. The main from-scratch experiments use $r=32$, which is 1.2 higher than $r=64$ at W2A16 and 0.4 to 0.5 lower at W4A16 and W4A4.

\textbf{Refresh interval.} Training converges for every $\tau$ from 50 to 2000. The best interval differs across settings, namely $\tau=400$ at W2A16 (19.8), $\tau=100$ at W4A16 (18.8), and $\tau=50$ at W4A4 (18.3). Within each setting, perplexity varies by about 1 over the whole range, and it shows no consistent trend with $\tau$. Refreshing more often than every 400 steps therefore does not give a consistent gain, so the SVD cost can be spread over hundreds of steps. We use $\tau=400$ for from-scratch QAT, which is within 0.9 of the best interval in every setting.

\begin{figure}[h]
    \centering
    \vspace{-0.18in}
    \includegraphics[width=0.86\columnwidth]{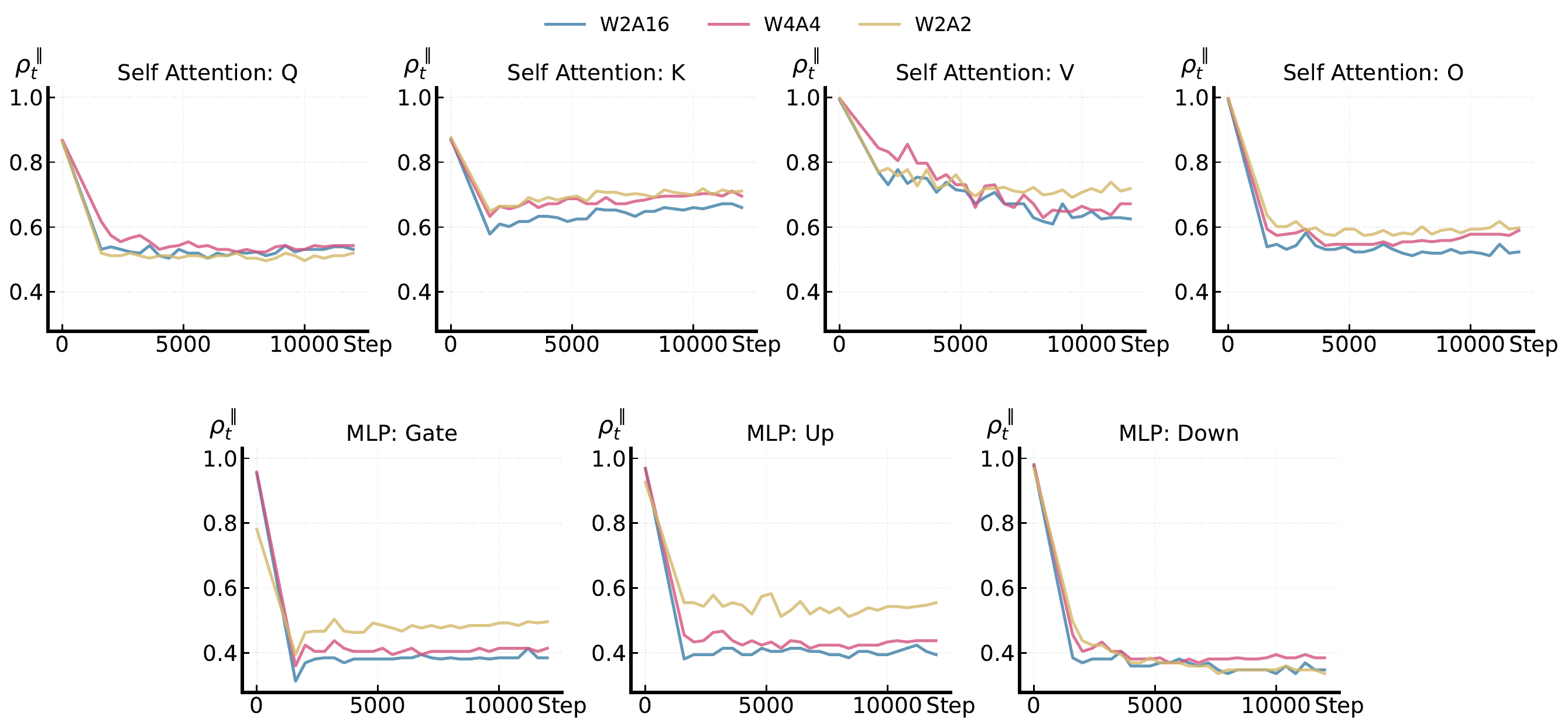}
    \vspace{-0.13in}
    \caption{Energy ratio of the principal subspace ($\rho_t^{\parallel}$) during from-scratch QAT on a 297M Llama-style model. The curves represent different weight matrices within a middle transformer layer across W2A16, W4A4, and W2A2 quantization configurations.}
    \vspace{-5px}
    \label{fig:energy_ratio}
\end{figure}

\vspace{-6px}
\subsection{Dynamics of Momentum Energy Distribution}
\label{sec:energy_dynamics}
We track the energy ratio of the principal subspace, $\rho_t^{\parallel}=\|M_t^{\parallel}\|_F/(\|M_t\|_F+\epsilon)$, for the seven weight matrices of a middle layer in the 297M model pretrained with ResQAdam. We record it only at refresh steps, right after the bases are recomputed from $M_t$. At these points $\rho_t^{\parallel}$ is at its highest, so any residual energy we observe is a lower bound on the energy that a projection-based optimizer discards during the following $\tau$ steps.

The result is shown in Figure~\ref{fig:energy_ratio}. $\rho_t^{\parallel}$ starts high, decreases, and settles at a level that depends on the matrix. 
It remains relatively high for attention K and V and is lowest for the MLP down projection, while the bit-width setting has a smaller effect. 
Even immediately after a refresh, a substantial share of the momentum lies outside $\mathcal{S}_t$, and projection-based optimizers discard this share.

Overall, the principal subspace consistently captures a meaningful but incomplete fraction of the momentum. This supports ResQAdam's selective-adaptivity design: Adam-style preconditioning is beneficial for the structured principal component, while a full-space residual correction remains necessary to preserve the substantial optimization signal outside the low-rank subspace.

\vspace{-0.07in}
\section{Conclusion}
\vspace{-0.05in}
In this paper, we proposed ResQAdam, a memory-efficient optimizer tailored for LLM QAT. We demonstrated that standard low-rank optimizers struggle in QAT because they discard important residual optimization signals outside the dominant principal subspace.
ResQAdam leveraged Adam within this subspace and adds a sign update from the residual.
Extensive experiments across from-scratch QAT and fine-tuning show that ResQAdam closely tracks AdamW's convergence behavior and largely preserves its final performance, while reducing persistent optimizer-state memory by 71\% to 74\%, respectively.

\subsection*{AI use statement}
In this work, we used generative AI tools for polishing the writing, correcting grammar, assisting with literature search and figure plotting, and locating code bugs. We have not used generative AI tools for developing the core ideas, methods, or contributions of this work. We have reviewed all AI-assisted work: polished text was checked for accuracy, retrieved literature was verified against the original sources, figures were checked against the experimental data, and suggested bug fixes were manually tested before adoption. We take responsibility for the final content of this work, including text, claims or artifacts produced with the aid of generative AI.

\subsection*{Ethics Statement}
This work focuses on methodological improvements and does not involve human subjects, personally identifiable information, or the collection of new data. All datasets and models used in our experiments are publicly available and were used in accordance with their respective licenses. We do not foresee any direct negative societal impact or ethical concerns arising from this work. The authors declare no conflicts of interest.

\subsection*{Reproducibility statement}
We have made efforts to ensure the reproducibility of this work. The details of proposed method are described in Section~\ref{resq}. Implementation details, including hyperparameters, training settings, and hardware configurations, are provided in Section~\ref{sec:exp_setup} and Appendix~\ref{app:experiments_cfg}. The datasets and models used in our experiments are publicly available. Our code is available at \url{https://github.com/LiqunMa/ResQAdam}.

\bibliography{iclr2027_conference}

@article{frantar2022gptq,
  title={Gptq: Accurate post-training quantization for generative pre-trained transformers},
  author={Frantar, Elias and Ashkboos, Saleh and Hoefler, Torsten and Alistarh, Dan},
  journal={arXiv preprint arXiv:2210.17323},
  year={2022}
}

@inproceedings{xiao2023smoothquant,
  title={Smoothquant: Accurate and efficient post-training quantization for large language models},
  author={Xiao, Guangxuan and Lin, Ji and Seznec, Mickael and Wu, Hao and Demouth, Julien and Han, Song},
  booktitle={International conference on machine learning},
  pages={38087--38099},
  year={2023},
  organization={PMLR}
}

@inproceedings{liu2024llmqat,
  title={Llm-qat: Data-free quantization aware training for large language models},
  author={Liu, Zechun and Oguz, Barlas and Zhao, Changsheng and Chang, Ernie and Stock, Pierre and Mehdad, Yashar and Shi, Yangyang and Krishnamoorthi, Raghuraman and Chandra, Vikas},
  booktitle={Findings of the Association for Computational Linguistics: ACL 2024},
  pages={467--484},
  year={2024}
}

@inproceedings{chen2025efficientqat,
  title={Efficientqat: Efficient quantization-aware training for large language models},
  author={Chen, Mengzhao and Shao, Wenqi and Xu, Peng and Wang, Jiahao and Gao, Peng and Zhang, Kaipeng and Luo, Ping},
  booktitle={Proceedings of the 63rd Annual Meeting of the Association for Computational Linguistics (Volume 1: Long Papers)},
  pages={10081--10100},
  year={2025}
}

@article{robert2025ldadam,
  title={Ldadam: Adaptive optimization from low-dimensional gradient statistics},
  author={Robert, Thomas and Safaryan, Mher and Modoranu, Ionut-Vlad and Alistarh, Dan},
  journal={arXiv preprint arXiv:2410.16103},
  year={2024}
}

@article{panferov2025quest,
  title={Quest: Stable training of llms with 1-bit weights and activations},
  author={Panferov, Andrei and Chen, Jiale and Tabesh, Soroush and Castro, Roberto L and Nikdan, Mahdi and Alistarh, Dan},
  journal={arXiv preprint arXiv:2502.05003},
  year={2025}
}

@article{liu2026paretoq,
  title={ParetoQ: Improving scaling laws in extremely low-bit LLM quantization},
  author={Liu, Zechun and Zhao, Changsheng and Huang, Hanxian and Chen, Sijia and Zhang, Jing and Zhao, Jiawei and Roy, Scott and Jin, Lisa and Xiong, Yunyang and Shi, Yangyang and others},
  journal={Advances in Neural Information Processing Systems},
  volume={38},
  pages={91311--91336},
  year={2026}
}

@article{zhao2024galore,
  title={Galore: Memory-efficient llm training by gradient low-rank projection},
  author={Zhao, Jiawei and Zhang, Zhenyu and Chen, Beidi and Wang, Zhangyang and Anandkumar, Anima and Tian, Yuandong},
  journal={arXiv preprint arXiv:2403.03507},
  year={2024}
}

@article{loshchilov2017decoupled,
  title={Decoupled weight decay regularization},
  author={Loshchilov, Ilya and Hutter, Frank},
  journal={arXiv preprint arXiv:1711.05101},
  year={2017}
}

@inproceedings{shazeer2018adafactor,
  title={Adafactor: Adaptive learning rates with sublinear memory cost},
  author={Shazeer, Noam and Stern, Mitchell},
  booktitle={International conference on machine learning},
  pages={4596--4604},
  year={2018},
  organization={PMLR}
}

@article{dettmers2023qlora,
  title={Qlora: Efficient finetuning of quantized llms},
  author={Dettmers, Tim and Pagnoni, Artidoro and Holtzman, Ari and Zettlemoyer, Luke},
  journal={Advances in neural information processing systems},
  volume={36},
  pages={10088--10115},
  year={2023}
}

@article{bengio2013estimating,
  title={Estimating or propagating gradients through stochastic neurons for conditional computation},
  author={Bengio, Yoshua and L{\'e}onard, Nicholas and Courville, Aaron},
  journal={arXiv preprint arXiv:1308.3432},
  year={2013}
}

@article{esser2019learned,
  title={Learned step size quantization},
  author={Esser, Steven K and McKinstry, Jeffrey L and Bablani, Deepika and Appuswamy, Rathinakumar and Modha, Dharmendra S},
  journal={arXiv preprint arXiv:1902.08153},
  year={2019}
}

@article{tabesh2025cage,
  title={CAGE: Curvature-Aware Gradient Estimation For Accurate Quantization-Aware Training},
  author={Tabesh, Soroush and Safaryan, Mher and Panferov, Andrei and Volkova, Alexandra and Alistarh, Dan},
  journal={arXiv preprint arXiv:2510.18784},
  year={2025}
}

@inproceedings{zellers2019hellaswag,
  title={Hellaswag: Can a machine really finish your sentence?},
  author={Zellers, Rowan and Holtzman, Ari and Bisk, Yonatan and Farhadi, Ali and Choi, Yejin},
  booktitle={Proceedings of the 57th annual meeting of the association for computational linguistics},
  pages={4791--4800},
  year={2019}
}

@inproceedings{bisk2020piqa,
  title={Piqa: Reasoning about physical commonsense in natural language},
  author={Bisk, Yonatan and Zellers, Rowan and Gao, Jianfeng and Choi, Yejin and others},
  booktitle={Proceedings of the AAAI conference on artificial intelligence},
  volume={34},
  number={05},
  pages={7432--7439},
  year={2020}
}

@article{clark2018think,
  title={Think you have solved question answering? try arc, the ai2 reasoning challenge},
  author={Clark, Peter and Cowhey, Isaac and Etzioni, Oren and Khot, Tushar and Sabharwal, Ashish and Schoenick, Carissa and Tafjord, Oyvind},
  journal={arXiv preprint arXiv:1803.05457},
  year={2018}
}

@article{sakaguchi2021winogrande,
  title={Winogrande: An adversarial winograd schema challenge at scale},
  author={Sakaguchi, Keisuke and Bras, Ronan Le and Bhagavatula, Chandra and Choi, Yejin},
  journal={Communications of the ACM},
  volume={64},
  number={9},
  pages={99--106},
  year={2021},
  publisher={ACM New York, NY, USA}
}

@article{hendryckstest2021,
      title={Measuring Massive Multitask Language Understanding},
      author={Dan Hendrycks and Collin Burns and Steven Basart and Andy Zou and Mantas Mazeika and Dawn Song and Jacob Steinhardt},
      journal={Proceedings of the International Conference on Learning Representations (ICLR)},
      year={2021}
    }

@article{wang2023bitnet,
  title={Bitnet: Scaling 1-bit transformers for large language models},
  author={Wang, Hongyu and Ma, Shuming and Dong, Li and Huang, Shaohan and Wang, Huaijie and Ma, Lingxiao and Yang, Fan and Wang, Ruiping and Wu, Yi and Wei, Furu},
  journal={arXiv preprint arXiv:2310.11453},
  year={2023}
}

@article{ma2024era,
  title={The era of 1-bit llms: All large language models are in 1.58 bits},
  author={Ma, Shuming and Wang, Hongyu and Ma, Lingxiao and Wang, Lei and Wang, Wenhui and Huang, Shaohan and Dong, Li and Wang, Ruiping and Xue, Jilong and Wei, Furu},
  journal={arXiv preprint arXiv:2402.17764},
  year={2024}
}

@article{choi2018pact,
  title={Pact: Parameterized clipping activation for quantized neural networks},
  author={Choi, Jungwook and Wang, Zhuo and Venkataramani, Swagath and Chuang, Pierce I-Jen and Srinivasan, Vijayalakshmi and Gopalakrishnan, Kailash},
  journal={arXiv preprint arXiv:1805.06085},
  year={2018}
}

@article{kim2023memory,
  title={Memory-efficient fine-tuning of compressed large language models via sub-4-bit integer quantization},
  author={Kim, Jeonghoon and Lee, Jung Hyun and Kim, Sungdong and Park, Joonsuk and Yoo, Kang Min and Kwon, Se Jung and Lee, Dongsoo},
  journal={Advances in Neural Information Processing Systems},
  volume={36},
  pages={36187--36207},
  year={2023}
}

@article{chen2026fira,
  title={Fira: Can we achieve full-rank training of llms under low-rank constraint?},
  author={Chen, Xi and Feng, Kaituo and Li, Changsheng and Lai, Xunhao and Yue, Xiangyu and Yuan, Ye and Wang, Guoren},
  journal={Advances in Neural Information Processing Systems},
  volume={38},
  pages={120680--120712},
  year={2026}
}

@article{zmushko2024frugal,
  title={FRUGAL: Memory-efficient optimization by reducing state overhead for scalable training},
  author={Zmushko, Philip and Beznosikov, Aleksandr and Tak{\'a}{\v{c}}, Martin and Horv{\'a}th, Samuel},
  journal={arXiv preprint arXiv:2411.07837},
  year={2024}
}

@misc{qwen3technicalreport,
      title={Qwen3 Technical Report}, 
      author={Qwen Team},
      year={2025},
      eprint={2505.09388},
      archivePrefix={arXiv},
      primaryClass={cs.CL},
      url={https://arxiv.org/abs/2505.09388}, 
}

@article{llama3modelcard,
  title={Llama 3 Model Card},
  author={AI@Meta},
  year={2024},
  url = {https://github.com/meta-llama/llama3/blob/main/MODEL_CARD.md}
}

@article{touvron2023llama,
  title={Llama 2: Open foundation and fine-tuned chat models},
  author={Touvron, Hugo and Martin, Louis and Stone, Kevin and Albert, Peter and Almahairi, Amjad and Babaei, Yasmine and Bashlykov, Nikolay and Batra, Soumya and Bhargava, Prajjwal and Bhosale, Shruti and others},
  journal={arXiv preprint arXiv:2307.09288},
  year={2023}
}

@software{together2023redpajama,
  author = {Together Computer},
  title = {RedPajama: An Open Source Recipe to Reproduce LLaMA training dataset},
  month = April,
  year = 2023,
  url = {https://github.com/togethercomputer/RedPajama-Data}
}
\bibliographystyle{iclr2027_conference}

\clearpage
\appendix

\section*{\Large{Appendix}}
\section{Estimating the Statistics of Projected Gradients}
\label{app:state_transport}

When the projection bases are refreshed, the coordinate representation of the
principal subspace changes from $(U_{t-1},V_{t-1})$ to $(U_t,V_t)$, where $t-1$
denotes the last step before the refresh. Since Adam-style preconditioning is
anisotropic, directly reusing the old principal moments $m_{t-1}^{\parallel}$ and
$v_{t-1}^{\parallel}$ in the new coordinate system would mix statistics computed
under different bases. To address this, we adopt the statistical state transport mechanism introduced by LDAdam~\citep{robert2025ldadam}, which estimates the first- and second-order statistics of the newly projected gradients. As noted in Section~\ref{sec:refresh_complexity}, only the compact principal states are transported. We treat this transport as a lightweight heuristic that maintains the continuity of the Adam-preconditioned principal update across refreshes.

Recall that ResQAdam projects the full-space gradient $G_t\in\mathbb{R}^{n\times m}$
into the $r\times r$ coordinate representation of the principal subspace as
\begin{equation}
    G_t^{\parallel} = U_t^\top G_t V_t \in \mathbb{R}^{r\times r}.
\end{equation}
Let the old and new bases define the transition matrices
\begin{equation}
    P_t = U_t^\top U_{t-1} \in \mathbb{R}^{r\times r},
    \qquad
    Q_t = V_{t-1}^\top V_t \in \mathbb{R}^{r\times r}.
\end{equation}
Under the rank-$r$ truncation induced by the old bases, the projection of any past gradient $G_k$ ($k < t$) into the new coordinate representation can be approximated by
\begin{equation}
    U_t^\top G_k V_t
    \approx
    \left(U_t^\top U_{t-1}\right)
    \left(U_{t-1}^\top G_k V_{t-1}\right)
    \left(V_{t-1}^\top V_t\right)
    =
    P_t G_k^{\parallel} Q_t .
\end{equation}
Because the optimizer states are essentially exponentially moving averages (EMAs) of these past gradients, this transformation forms the basis for transporting their statistics.

\paragraph{First-order statistic.}
The transport must be performed on the \textit{bias-corrected} statistical moments rather than the raw EMAs. This ensures that the first and second moments are on the same true scale when estimating the variance. Because the first moment is linear in the gradient, the transported first-order statistic is obtained by applying the coordinate rotation to the bias-corrected old first moment:
\begin{equation}
    \widehat{m}_{t-\frac{1}{2}}^{\parallel}
    =
    P_t \widehat{m}_{t-1}^{\parallel} Q_t .
\end{equation}
Equivalently, the raw transported first moment is
\begin{equation}
    m_{t-\frac{1}{2}}^{\parallel}
    =
    \left(1-\beta_1^{\,t-1}\right)
    \widehat{m}_{t-\frac{1}{2}}^{\parallel} .
\end{equation}

\paragraph{Second-order statistic.}
The second moment $v^{\parallel}$ is an element-wise squared-gradient EMA rather than a true covariance matrix, so it is not linear under a change of coordinates. For each new coordinate $(a,b)$, the corresponding projected gradient is approximated as
\begin{equation}
    \left(G_k^{\parallel,\mathrm{new}}\right)_{ab}
    \approx
    \sum_{i=1}^{r}\sum_{j=1}^{r}
    (P_t)_{ai} (Q_t)_{jb}
    \left(G_k^{\parallel,\mathrm{old}}\right)_{ij}.
\end{equation}
Assuming coordinate-wise independence as a heuristic, its second-order statistic can be estimated by transporting the variance via the squared transition matrices and re-adding the squared transported mean:
\begin{align}
    \left(\widehat{v}_{t-\frac{1}{2}}^{\parallel}\right)_{ab}
    &\approx
    \sum_{i=1}^{r}\sum_{j=1}^{r}
    (P_t)_{ai}^{2}(Q_t)_{jb}^{2}
    \left[
        \left(\widehat{v}_{t-1}^{\parallel}\right)_{ij}
        -
        \left(\widehat{m}_{t-1}^{\parallel}\right)_{ij}^{2}
    \right]
    +
    \left(\widehat{m}_{t-\frac{1}{2}}^{\parallel}\right)_{ab}^2 .
    \label{eq:resqadam_second_moment_transport}
\end{align}
The first term estimates the transformed coordinate-wise variance, while the second term accounts for the squared transformed mean. In practice, to avoid numerical artifacts caused by this independence approximation, we clip the transported second-order estimate to be non-negative:
\begin{equation}
    \widehat{v}_{t-\frac{1}{2}}^{\parallel}
    \leftarrow
    \max\left(
        \widehat{v}_{t-\frac{1}{2}}^{\parallel}, 0
    \right).
\end{equation}
The raw transported second moment is then recovered by reapplying the bias term:
\begin{equation}
    v_{t-\frac{1}{2}}^{\parallel}
    =
    \left(1-\beta_2^{\,t-1}\right)
    \widehat{v}_{t-\frac{1}{2}}^{\parallel} .
\end{equation}

After this transport step, ResQAdam continues the principal Adam update in the refreshed coordinate representation:
\begin{align}
    G_t^{\parallel} &= U_t^\top G_t V_t, \\
    m_t^{\parallel}
    &=
    \beta_1 m_{t-\frac{1}{2}}^{\parallel}
    +
    (1-\beta_1)G_t^{\parallel}, \\
    v_t^{\parallel}
    &=
    \beta_2 v_{t-\frac{1}{2}}^{\parallel}
    +
    (1-\beta_2)
    \left(G_t^{\parallel}\odot G_t^{\parallel}\right).
\end{align}

\paragraph{Difference in Application Scope.}
While the statistical transport equations strictly follow LDAdam~\citep{robert2025ldadam}, our application differs in scope. 
LDAdam relies on this transport to perform adaptive optimization entirely within the compressed principal subspace, using generalized error feedback to reintroduce projection errors. 
In contrast, ResQAdam restricts this transport solely to the compact principal states $m_t^{\parallel},v_t^{\parallel}\in\mathbb{R}^{r\times r}$. 
We maintain a dense momentum $M_t$ in the original parameter space (stored in BF16) and explicitly preserve the discarded orthogonal residual $M_t^{\perp}$ through a full-space sign correction. 
Thus, our use of the projected-gradient statistic estimation serves purely as a continuity mechanism for the principal update, deliberately avoiding the compression of residual QAT compensations into LDAdam-style dense second-moment tracking.

\section{Training loss trajectories on 200M-parameter LLM}
\label{app:200m}
\begin{figure}[t]
    \centering
    \includegraphics[width=0.9\columnwidth]{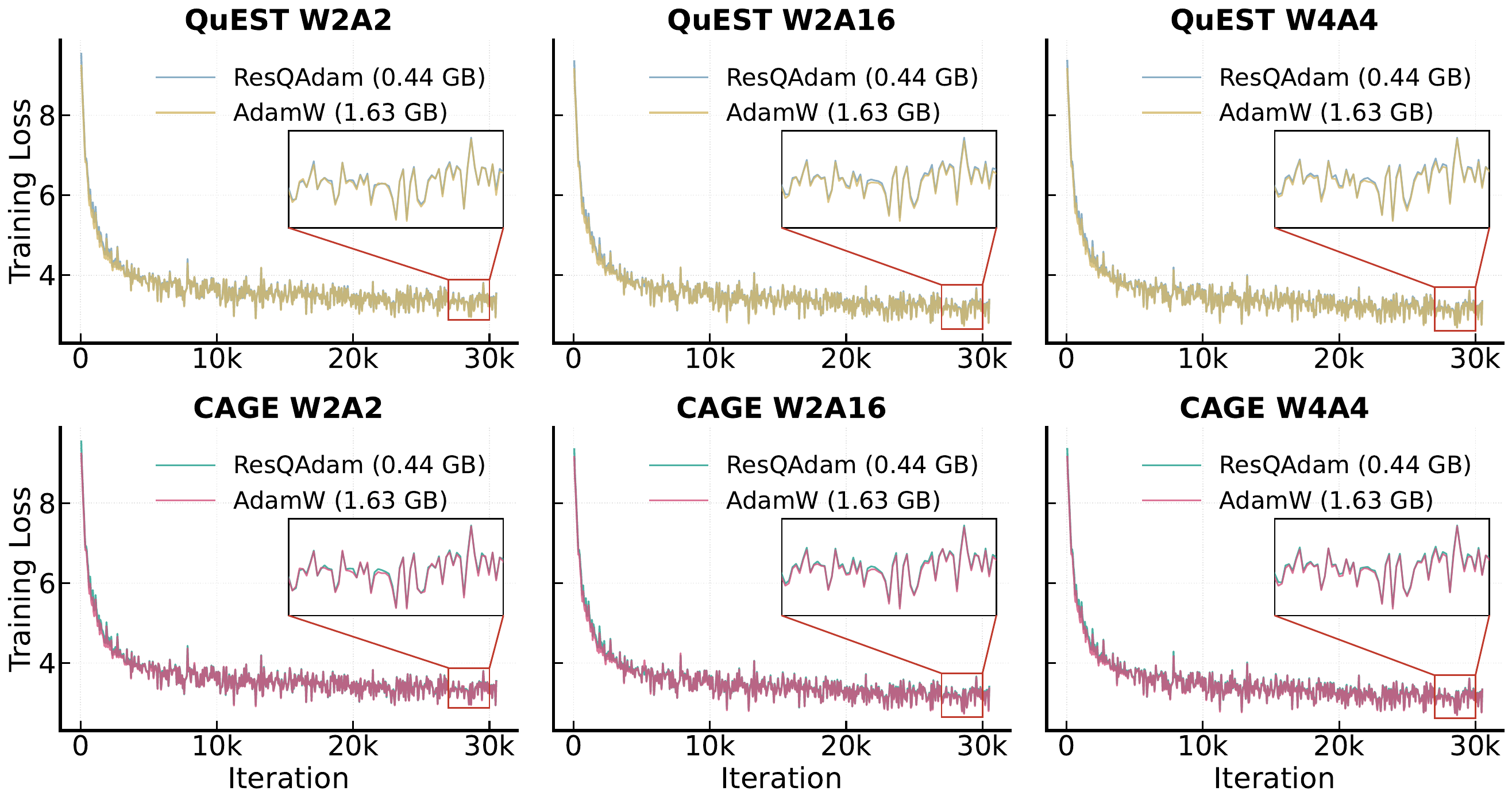}
    \caption{Training loss trajectories of ResQAdam and AdamW during from-scratch QAT under various quantization configurations on the 200M-parameter LLM. The values following the optimizer names in the legend indicate the average memory footprint of the optimizer states.}
    \label{fig:loss_200m}
\end{figure}

Figure~\ref{fig:loss_200m} shows the optimization trajectory and convergence alignment during training for a 200M-parameter LLM, comparing ResQAdam and AdamW under two from-scratch QAT methods, QuEST and CAGE. For the 200M-scale model, ResQAdam exhibits a training-loss trajectory that closely tracks that of AdamW.

\section{QAT Frameworks and Experimental Configurations}
\label{app:experiments_cfg}
We evaluate ResQAdam across both from-scratch QAT and post-pretraining QAT fine-tuning, covering substantially different optimization objectives, training schemes, and quantization settings.
\subsection{From-Scratch QAT}
We consider two recent low-precision pretraining frameworks.
\paragraph{QuEST.}
QuEST~\citep{panferov2025quest} applies Hadamard transformations to weights and activations before quantization and uses a trust-based gradient estimator to reduce gradient bias from large quantization errors. We replace its original optimizer with ResQAdam while keeping the remaining training pipeline unchanged.
\paragraph{CAGE.}
CAGE~\citep{tabesh2025cage} augments the STE with a curvature-aware correction derived from a multi-objective formulation that jointly considers task loss and quantization error. ResQAdam is integrated without modifying its quantization or gradient-correction scheme.
For both frameworks, we pretrain Llama-style models with 200M and 430M parameters on C4 under W2A2, W2A16, and W4A4 quantization. The training-token budget is set to $40\times$ the model size.
\subsection{QAT Fine-Tuning}
We further evaluate ResQAdam under three QAT fine-tuning frameworks with different optimization objectives and data sources.
\paragraph{Basic QAT.}
Our Basic QAT baseline performs end-to-end next-token training with learnable asymmetric quantization, where quantization scales are optimized following LSQ~\citep{esser2019learned}. All parameters are jointly fine-tuned on a 2B-token subset of RedPajama~\citep{together2023redpajama}.
\paragraph{EfficientQAT (Block-AP).}
EfficientQAT~\citep{chen2025efficientqat} first performs block-wise all-parameter training (Block-AP), followed by an E2E-QP stage that updates only quantization parameters. Since ResQAdam operates on weight-matrix updates, we evaluate it only in the Block-AP stage.
\paragraph{LLM-QAT.}
LLM-QAT~\citep{liu2024llmqat} uses data-free knowledge distillation, generating training sequences from the full-precision teacher and optimizing the quantized student to match it. It jointly quantizes weights, activations, and the KV cache using symmetric MinMax quantization.
For fine-tuning experiments, we use Llama 3.2-1B~\citep{llama3modelcard}, Qwen 3-4B-Base~\citep{qwen3technicalreport}, and Llama 2-7B~\citep{touvron2023llama} under multiple low-bit quantization configurations. Exact configurations are reported with the corresponding experimental results.
\end{document}